# SIDME: Self-supervised Image Demoiréing via Masked Encoder-Decoder Reconstruction

**Xia Wang,** [a, *] **Haiyang Sun,** [b] **Tiantian Cao,** [a] **Yueying Sun,** [c] **Min Feng**[d]

[a]College of Artificial Intelligence, Beijing University of Posts and Telecommunications, Beijing 100876, China
[b]College of Electronic Information and Electrical Engineering, Shanghai Jiaotong University, Shanghai 200030, China
[c]College of Information, Shanghai Ocean University, Shanghai 201306, China
[d]College of Computer, Beijing University of Posts and Telecommunications, Beijing 100876, China

**Abstract**. Moiré patterns, resulting from aliasing between object light signals and camera sampling frequencies, often degrade image quality during capture. Traditional demoiréing methods have generally treated images as a whole for processing and training, neglecting the unique signal characteristics of different color channels. Moreover, the randomness and variability of moiré pattern generation pose challenges to the robustness of existing methods when applied to real-world data. To address these issues, this paper presents SIDME (Self-supervised Image Demoiréing via Masked Encoder-Decoder Reconstruction), a novel model designed to generate high-quality visual images by effectively processing moiré patterns. SIDME combines a masked encoder-decoder architecture with self-supervised learning, allowing the model to reconstruct images using the inherent properties of camera sampling frequencies. A key innovation is the random masked image reconstructor, which utilizes an encoder-decoder structure to handle the reconstruction task. Furthermore, since the green channel in camera sampling has a higher sampling frequency compared to red and blue channels, a specialized self-supervised loss function is designed to improve the training efficiency and effectiveness. To ensure the generalization ability of the model, a self-supervised moiré image generation method has been developed to produce a dataset that closely mimics real-world conditions. Extensive experiments demonstrate that SIDME outperforms existing methods in processing real moiré pattern data, showing its superior generalization performance and robustness.

**\*** Corresponding Author**,** E-mail: xiaonini@bupt.edu.cn

## 1 Introduction

Moiré patterns arise from frequency mixing between the sampling frequency of the camera and the periodic light signals emitted by the subject. To elucidate the root cause and mechanism of moiré patterns, we take the example of photographing an LED screen, whose light-emitting units of an LED screen exhibit a certain periodic structure. When these signals are captured by an image sensor equipped with a Bayer color filter array, frequency aliasing occurs if the sampling frequency of the sensor is close but not identical to the frequency of the emitted light signals. This frequency misalignment leads to the formation of moiré patterns. Moiré patterns exhibit significant variations in shape, color, and frequency, making it difficult to distinguish them from the intrinsic textures of the captured image. Additionally, the manifestation of moiré is characterized by randomness, and variations in moiré can be observed from image to image as well as within different regions of the same image. These multifaceted and unpredictable characteristics present a great challenge to image processing algorithms designed to mitigate or eliminate moiré patterns while preserving the integrity of the original image content.

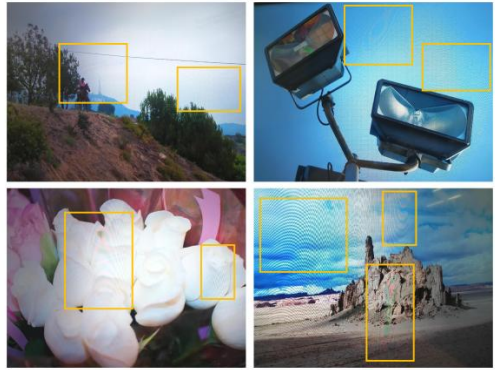

**Fig. 1** Exemplary image of authentic moiré patterns.

To address this challenge, previous research has mainly focused on processing and training images as a whole. As shown in Fig. 2, the Bayer sampling array RGGB in sensors of mainstream imaging devices such as cameras and smartphones, captures color images by organizing R (red), double G (green), and B (blue) pixels in a defined pattern. It is important to note that the green channel, endowed with double the sampling frequency of red and blue channels, experiences substantially less aliasing-induced information loss. Consequently, as illustrated in Fig. 3, the green channel shows significantly fewer moiré patterns. However, prior studies have not fully utilized this unique advantage, resulting in only partial mitigation of moiré noise and limited robustness in real-world data scenarios. Moreover, the inherent randomness and complexity of moiré patterns pose a huge obstacle to existing methods, thereby preventing the achievement of optimal demoiréing outcomes.

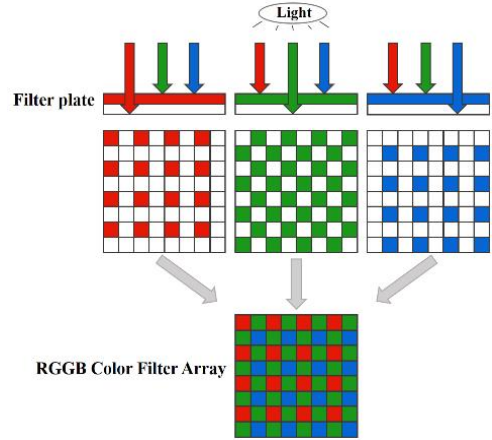

**Fig. 2** Example of Bayer color filter array.

To tackle the above challenges, we propose SIDME (Self-supervised Image Demoiréing via Masked Encoder-Decoder Reconstruction), an innovative model designed to enhance image quality by reducing moiré patterns. SIDME integrates a masked encoder-decoder reconstruction module with a self-supervised learning framework aimed at generating high-quality visual images, overcoming the limitations of existing algorithms in processing real-world moiré data. Within the SIDME framework, a random masked image reconstructor is designed to utilize an encoder-decoder architecture for image reconstruction. Given the inherent randomness and complexity of moiré noise, which pose significant challenges in designing effective demoiréing algorithms, a random image mask is introduced to counteract this randomness and enhance the robustness of the algorithm when dealing with real-world data. By applying a specific percentage of the mask to moiré images, random patches are created within the image, and the missing patches are reconstructed using the encoder-decoder structure. Furthermore, the unique characteristics of camera sampling are fully exploited, in particular, the double sampling frequency of the green channel compared to the red and blue channels. This characteristic is leveraged to reduce aliasing-induced information loss and further enhance the quality of the reconstructed images.

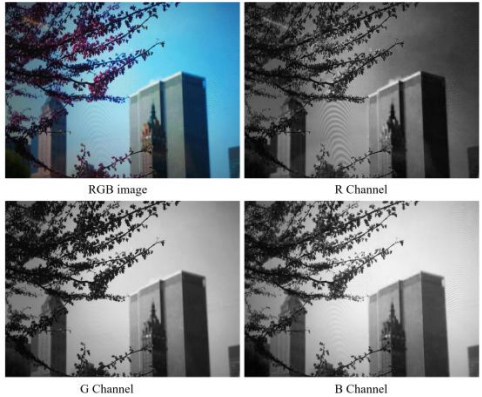

**Fig. 3** RGB channels of authentic moiré images.

Notably, we develop a specialized self-supervised loss function that leverages the intrinsic features of the image to guide model training. This innovation further strengthens the model's ability to combat the randomness of moiré patterns, enabling it to better remove moiré noise while preserving normal image textures and enhancing its generalization capabilities when processing real-world data. To further enhance training efficiency for demoiréing networks, a moiré pattern generation approach is developed. Compared to existing public moiré datasets, our method generates image data that closely resembles authentic moiré patterns while ensuring consistency in tone and brightness between moiré image and corresponding clean images. This provides more effective training data and enhances the model's generalization capability. Extensive experimental validation confirms that the SIDME model exhibits robust generalization performance when processing real-world moiré data, outperforming existing methods and demonstrating significant advantages in various scenarios.

In summary, the primary contributions of this paper are threefold:

1. We innovatively propose the SIDME model, which utilizes a random mask encoding-decoding architecture to reconstruct image patches. By integrating self-supervised techniques, the model effectively addresses the inherent randomness and disorder of moiré patterns, thereby enhancing its robustness and generalization capacity when processing real-world moiré data.

2. By exploiting the doubled green channel sampling in the Bayer RGGB array, we devise a self-supervised loss function that leverages intrinsic image information to enhance the model's adaptive demoiréing capability, particularly for real-world data.

3. A novel method has been developed for generating image data that incorporates realistic moiré patterns while ensuring consistent tone and brightness between moiré and non-moiré images, thereby providing more effective data support for model training.

## 2 Related Works

### 2.1 Moiré Image Dataset

Existing datasets for moiré pattern model training primarily consist of two categories: real datasets and synthetic datasets. Real datasets, exemplified by [8-12], involve capturing screen images exhibiting moiré patterns using cameras and utilizing screenshots as the corresponding original images devoid of moiré patterns. However, this approach introduces notable discrepancies in brightness and hue between the image pairs, along with considerable limitations in alignment accuracy. Synthetic datasets, such as those in [13-16], obtain moiré images through simulated synthesis and designate the initial images as the corresponding moiré-free references. Nevertheless, these datasets inadequately replicate real-world moiré data and overlook significant differences in brightness and hue between the synthetic moiré images and ground truth, thereby failing to furnish effective training data crucial for model training.

### 2.2 Image Mask Encoder

Encoders constitute a classic method for learning input representations, wherein input encodings are mapped to latent feature representations, and decoders are subsequently employed for feature decoding and reconstruction. Masked encoding techniques address input data containing patches due to masking disruption, a methodology successfully applied in both natural language processing (NLP) and image processing domains. For instance, BERT [1] and GPT [2, 3] in NLP preserve segments of the input language sequence and train models to predict the missing parts, demonstrating robust generalization capabilities in practical scenarios and various downstream tasks [2]. In the field of image processing, [4] innovatively treated masking as a noise type, while iGPT [5] predicts unknown pixels by manipulating pixel sequences. ViT [6] investigated masked patch prediction for self-supervised learning, and MAE [7] designed a lightweight encoder-decoder architecture to predict missing image patches.

*2.3 Image Demoiréing*

The existing methods for eliminating moiré patterns in images can be broadly categorized into traditional methods and deep learning algorithms. Traditional methods primarily focus on modifying camera optical components, such as adding an antialiasing low-pass filter in front of the lens to reduce periodic variations in the optical signal [17], alongside interpolation-based methods [18], filtering techniques [19], and layer decomposition [20]. While effective for removing regular and uniform moiré noise, these traditional methods exhibit limited applicability in scenarios involving large and complex moiré patterns. Consequently, they often result in compromised image details, such as blurring or overshoot artifacts. In contrast, deep learning-based demoiréing algorithms have proven more effective in practice.

In terms of algorithm design, these approaches primarily encompass multi-scale networks, global-to-local networks, and transform-domain networks. Multiscale networks include those with UNet architectures or encoder-decoder structures [8,9, 21-27], global-to-local networks exemplified by [11], and transform-domain networks such as those combining pixel and frequency domains [28] and wavelet-domain-based networks [12]. In a recent study, Niu et al. [29] proposed a multi-stage progressive processing method for demoiréing, where each stage focuses on moiré with different frequency intensities to reconstruct natural texture through multi-scale features. However, this method may introduce unnatural artifacts or transitional synthetic details, and is not applicable to realistic as well as complex scenes. Xiao et al. [30] segmented the ultra-high-resolution image into overlapping blocks to be processed separately. In addition, a pyramidal network structure is used to extract moiré features from different scales and gradually refine the demoiré results. Meanwhile, shallow and deep features are combined to enhance texture recovery. This method may lead to boundary artifacts, which can mistakenly damage the real high-frequency texture and cause local blurring.

In addition, some other scholars approach the problem from the perspective of frequency domain. Zhong et al. [31] proposed a training method based on unpaired real data, which performs adversarial training in both spatial and frequency domains simultaneously. But this method relies on data distribution alignment, which leads to excessive smoothing or loss of details in the image. Qi et al. [32] first converted the image to the frequency domain for adaptive peak suppression, and then enhanced the visual quality through spatial domain operation. Unfortunately, the filtering inevitably mistreats high-frequency details. Yang et al. [33]

performed moiré suppression in the frequency domain and texture restoration in the null domain. Additionally, they have designed a color correction module and introduced a dynamic weight allocation module. However, this method has limited effect on strongly nonperiodic demoiré removal.

In summary, the generalization capability of existing deep learning algorithms remains suboptimal. These models typically process images holistically during training and learning, which limits their robustness when applied to real-world data. Specifically, they often struggle with diverse and complex moiré patterns that deviate from the training distribution. Moreover, these methods fail to fully account for the sampling characteristics inherent in images, neglecting the exploitation of key intrinsic features that could otherwise enhance performance. This oversight further underscores the need for more robust and adaptive strategies in moiré pattern removal.

*2.4 Training Loss*

Existing image demoiréing algorithms optimize training using the original or deformed loss functions tailored to pixel-wise differences, perceptual quality, perceptual realism, and high-frequency details: L1 or L2 losses [21, 27] measure pixel-wise differences between the ground truth and reconstructed images, with L1 improving robustness and L2 ensuring smoother gradients; VGG loss [11, 24, 26, 28] captures perceptual similarities using deep features from pre-trained networks, enhancing visual quality by focusing on high-level structural and semantic features; adversarial loss [9] improves perceptual realism by making the output indistinguishable from real, moiré-free images through adversarial training; and high-frequency domain loss [20] focuses on recovering fine textures and details, minimizing the impact of moiré artifacts. In summary, the design of training loss functions in image demoiréing algorithms is critical to achieving high-quality results.

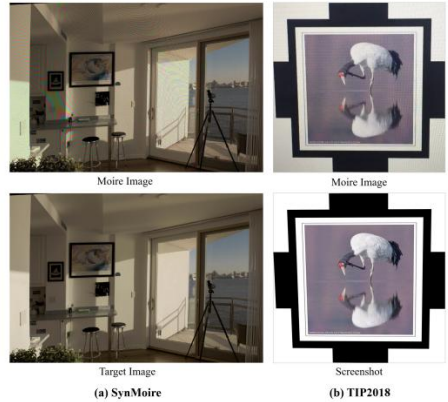

**Fig. 4** Example of the Dataset.

## 3 Datasets Construction

In constructing the synthetic moiré dataset, we utilize the fiveK dataset [34] as the base image and optimize the moiré synthesis methodology with reference to [14]. The specific steps are outlined as follows: Firstly, the original image is resampled to an RGB sub-pixel arrangement format. Subsequently, a random projection transformation is applied to simulate the effects of various camera shooting angles. Next, the radial distortion function is employed to mimic lens distortion phenomena. Afterwards, a Gaussian filter is used to simulate the camera's anti-aliasing filtering process. Following this, the image is resampled using a Bayer color filter array (CFA) to emulate the camera sensor's data acquisition process. Subsequently, the Bayer image undergoes demosaicing, and JPEG compression noise with a quantization factor of 0.5 is introduced to produce a composite image exhibiting moiré patterns.

To ensure the accuracy of the obtained data in terms of color tone and brightness, we further refine the synthesized moiré image based on the original image. Specifically, the image is converted to the YCbCr color space, and the Cb and Cr channels are adjusted according to the average values calculated from the 0–255-pixel positions in the original image. These adjusted average values are then assigned to the corresponding pixels in the synthesized image. Finally, the image is converted back to the RGB color space, resulting in the final image pair. We name this dataset SynMoiré, which comprises 1762 sets of images, each consisting of a moiré image and its corresponding clean image. Among these, 1234 sets are designated for training, with the remaining 528 sets (264 sets each for validation and testing) are utilized for assessment purposes. Example images are presented in Fig. 4, and compared to the widely used public dataset TIP2018 [8], the moiré effect in SynMoiré is more akin to real-world scenarios. Apart from exhibiting moiré noise, the differences between image pairs in SynMoiré are minimal, thereby providing robust support for effective model training.

Additionally, to assess the generalization capability of the model on real data, we capture an additional 231 real moiré images for testing purposes.

## 4 The Proposed Method

This paper presents SIDME, a method designed to address the issue of moiré patterns in images. The proposed model generates image patches with a specific mask ratio applied, and employs an encoder-decoder structure to reconstruct the images containing these patches. Additionally, considering the characteristics of image sampling, a self-supervised loss function is designed and combined with commonly used image processing loss functions for training the model. The overall framework of SIDME is illustrated in Fig. 5 and is further elaborated in the subsequent sections.

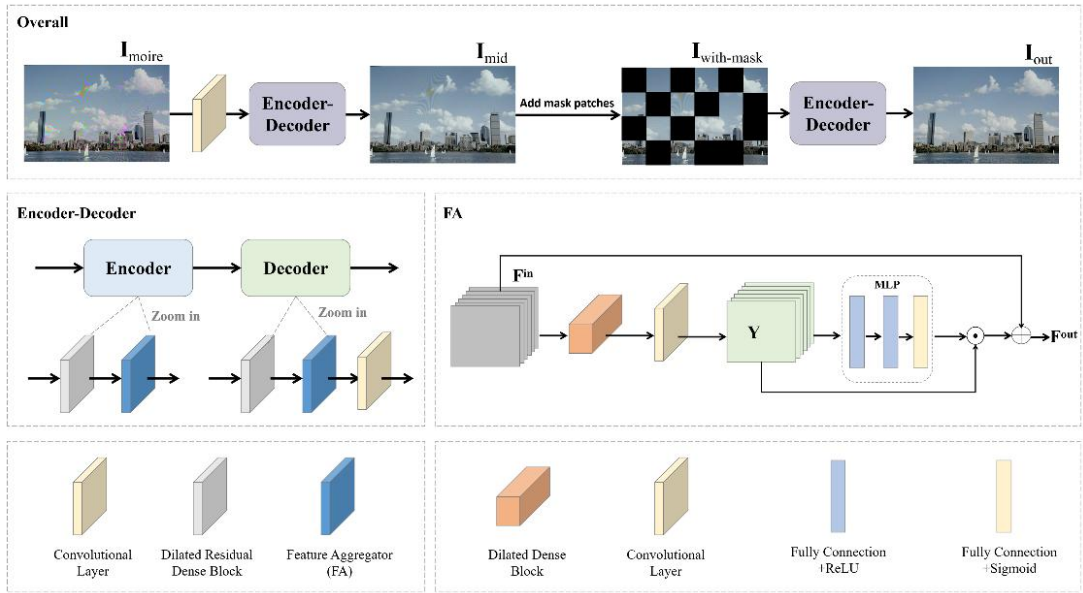

**Fig. 5** The structure of the SIDME and the proposed feature aggregator (FA).

### 4.1 Pipeline

As illustrated in Fig. 5, for the input moiré image $I_{moiré}$, an initial preprocessing step is conducted to augment the receptive field. Inspired by [9], we devised a demoiréing encoder-decoder architecture. This architecture processes the preprocessed image for preliminary demoiréing, yielding an intermediate result $\mathbf{I}_{mid}$. To further enhance the demoiréing effect, we innovatively introduced the mask patch, generating a masked image $\mathbf{I}_{with-mask}$ and performing mask-based encoder-decoder reconstruction to obtain the final moiré free image $\mathbf{I}_{out}$. In this pipeline, preprocessing involves using a 3×3 convolutional layer to deeply extract low-level features from the image. The encoder-decoder internally incorporates a dilated residual dense block (DRDB) [9] for refining input features, alongside a feature aggregator (FA). The decoder section appends an additional 3×3 convolutional layer at the end to accomplish feature mapping.

**Feature Aggregator** For the input image features $\mathbf{F}^{in}$, the Feature Aggregator (FA) proceeds with the following processing: Initially, $\mathbf{F}^{in}$ is input into a dilated dense block, which leverages multiple densely connected layers to deeply mine correlations and complementarity among features, thereby augmenting their representational power. Subsequently, the refined features traverse a 1×1 convolutional layer, which effectively reduces dimensionality and fine-tunes the features, yielding a feature map $\mathbf{Y}$ with enhanced information density. To fully harness the information within $\mathbf{Y}$, we devise a sophisticated fusion strategy that integrates $\mathbf{Y}$ with the original input features $\mathbf{F}^{in}$. This fusion is not a simplistic superposition but rather employs a dynamic weight allocation mechanism, with the fusion weights for $\mathbf{Y}$ being dynamically learned by a Multi-Layer Perceptron (MLP) module. The MLP module comprises three fully connected layers. By employing nonlinear activation functions and regularization techniques, these layers ensure the precision and robustness of the learned weights. After the fusion process, an output feature vector $\mathbf{F}^{out}$ is derived, encapsulating a rich set of feature information. Following the fusion step, an output feature $\mathbf{F}^{out}$ is obtained, which encapsulates a diverse array of feature information.

**Image Reconstruction Using Mask Patches** After obtaining the intermediary result $\mathbf{I}_{mid}$ through initial demoiréing from $\mathbf{I}_{moiré}$, we apply a masking technique to $\mathbf{I}_{mid}$. The mask consists of random image patches with a ratio of 0.6, whose positions, sizes, and shapes are randomly generated to ensure diversity in the image data across different processing instance. This process yields a masked image containing random patches, which is denoted as $\mathbf{I}_{with-mask}$. To further enhance the image quality, we perform encoding-decoding reconstruction on $\mathbf{I}_{with-mask}$. The adoption of image masking technology is intricately linked to the formation mechanism of moiré patterns. Moiré patterns are caused by the aliasing effect between the camera's sampling frequency and the signal frequency emitted by the photographed object. This aliasing effect, characterized by its randomness and unpredictability, generates a series of complex and unforeseeable interference stripes in the image, significantly impairing image quality and clarity. Consequently, it poses greater requirements on the robustness and generalization performance of image processing models. To address this challenge, we introduce the concept of random image patches, thereby enhancing the diversity and complexity of the image data. This approach enables the model to learn more about image structure and texture. Consequently, when processing real-world image data, the model can more effectively adapt to various complex scenarios, thus enhancing its robustness and generalization capabilities.

*4.2 Loss*

To train the image demoiréing model, we have devised a multidimensional loss function that incorporates basic loss, self-supervised loss, perceptual loss, and edge loss, as detailed below.

**Basic loss** In prior image processing studies [35,36], L1 loss has been demonstrated to outperform L2 loss in image restoration. To ensure that the model's output closely approximates the target moiré-free image, we adopt L1 loss as the basic loss function, i.e

$$L_{basic} = \left\| \mathbf{I}_{out} - \mathbf{I}_{target} \right\|_1 \tag{1}$$

where $\mathbf{I}_{out}$ is the final output of the network and $\mathbf{I}_{target}$ is the target image.

**Self-supervised loss** To enhance model training, we leverage the characteristic of image sampling, namely, that the sampling frequency of the green channel is double that of the blue and red channels, resulting in significantly weaker moiré patterns in the green channel compared to the blue and red channels. Based on this characteristic, we design a self-supervised loss function, utilizing the information from the green channel to super-vise the information in the red and blue channels. The self-supervised loss is computed using the intermediate result $\mathbf{I}_{mid}$, rather than the final result. The use of $\mathbf{I}_{mid}$ is aimed at correcting the information in the red and blue channels prior to mask encoding-decoding reconstruction, thereby improving the final demoiréing effect, i.e.

$$L_{supervised} = \left( \left\| \mathbf{I}_{mid}^{g} - \mathbf{I}_{mid}^{r} \right\| - \left\| \mathbf{I}_{mid}^{g} - \mathbf{I}_{mid}^{b} \right\| \right)_1 / 2 \tag{2}$$

where $\mathbf{I}_r^{mid}$, $\mathbf{I}_g^{mid}$, $\mathbf{I}_b^{mid}$ represent the red, green, and blue channels of $\mathbf{I}_{mid}$, respectively.

**Perceptual loss** Inspired by [9], to ensure the perceptual quality and subjective effectiveness of demoiréing results, particularly in regions with diverse semantic features, we design the perceptual loss. This loss is computed based on features and information extracted from images using the pre-trained VGG16 network [37], i.e.

$$L_{perceptual} = \left\| \Psi(\mathbf{I}_{out}) - \Psi(\mathbf{I}_{target}) \right\|_1 \quad (3)$$

where $\mathbf{I}_{out}$ is the final output of the network and $\mathbf{I}_{target}$ is the target image.

**Edge loss** To further enhance the refinement of high-frequency details in the processed results, this loss function is designed to optimize the detail representation of the image by quantifying the preservation of edge information. We utilize the Canny edge detector to extract the edges of the image and further compute the edge loss accordingly, i.e.

$$L_{edge} = \left\| Canny(\mathbf{I}_{out}) - Canny(\mathbf{I}_{target}) \right\|_1 \quad (4)$$

where $\mathbf{I}_{out}$ represents the model output, $\mathbf{I}_{target}$ represents the ground truth, *Canny* represents the Canny edge detector.

In conclusion, the overall loss function formulated for training the network is defined as follows:

$$L = L_{basic} + \lambda_s L_{supervised} + \lambda_p L_{perceptual} + \lambda_{edge} L_{edge} \quad (5)$$

where $\lambda_s$, $\lambda_p$, and $\lambda_{edge}$ are hyper-parameters that control the relative importance. In this paper, based on previous research experience in the field of image restoration and specific experimental results, the values of $\lambda_s$, $\lambda_p$, and $\lambda_{edge}$ are set to 10, 1, and 0.1, respectively.

## 5 Experiment and Results

### 5.1 Experiment Details

Our algorithm is implemented using PyTorch, and all experiments are conducted on NVIDIA RTX 3090 GPU. During model training, we set the batch size to 1 and optimize using the Adam optimizer [38] with parameters β1 = 0.9 and β2 = 0.999. The learning rate is initialized at 0.0003 and is subject to linear decay, halving every 30,000 iterations. The model is trained for 100 iterations. The image masking ratio is set to 0.6 To comprehensively and objectively evaluate image quality, we select Peak Signal-to-Noise Ratio (PSNR), Structural Similarity Index (SSIM), and Learned Perceptual Image Patch Similarity (LPIPS) as metrics, referencing [8, 12, 21].

## 5.2 Comparisons with State-of-the-art Methods

We evaluate the proposed SIDME against the current state-of-the-art demoiréing algorithms and image restoration algorithms using our self-constructed dataset SynMoiré, public dataset TIP2018 and real-world data. The benchmarked algorithms include demoiréing algorithms such as DMCNN [8], MopNet[21], MBCNN[24], WDNet[12], ESDNet[9], as well as image restoration algorithms such as P-BiC[30], Uformer[39], SwinIR[40], Wang[41]. All methods are trained on SynMoiré and tested on SynMoiré, TIP2018 and real-world image data to ensure a comprehensive and fair comparison.

The experimental results, as depicted in Fig. 6, Fig. 7, Tab. 1 and Tab.2, clearly demonstrate the superior performance of our model across various evaluation metrics. On the SynMoiré dataset, our model SIDME achieves remarkable results in terms of PSNR, SSIM, and LPIPS. The PSNR values for SIDME stand at 34.94, significantly exceeding those of P-BIC (32.98), ESDNet (32.65), and SwinIR (32.16). Similarly, in SSIM, our model achieves a value of 0.94, outperforming MBCNN (0.89), Uformer (0.9), ESDNet (0.93), and Wang (0.90). Additionally, SIDME's LPIPS score of 0.08 is notably lower than those of DMCNN (0.2), MopNet (0.16), WDNet (0.1), and SwinIR (0.16), indicating superior perceptual quality. While the LPIPS score is marginally inferior to that of ESDNet (0.0750), the pronounced advantages in PSNR and SSIM metrics indicate that our approach maintains superior overall performance. And on the SynMoiré dataset, our model SIDME achieves optimal results in terms of PSNR, SSIM and LPIPS, significantly outperforming other models.

Subsequently, a comprehensive evaluation of the SIDME and various other advanced methods is conducted on the TIP2018 dataset. The SIDME achieved a PSNR value of 30.33, which is close to the high-performing P-BIC (30.56) and significantly higher than methods like DMCNN (26.77) and SwinIR (26.15). This indicates that SIDME exhibits excellent performance in suppressing image noise while preserving image information. Besides, our model also performed well in LPIPS, scored only 0.14, which is much lower than other methods. This indicates that the images generated by SIDME are highly similar to real images in visual quality, aligning more closely with human visual perception, thus demonstrating the generalization capability of the method.

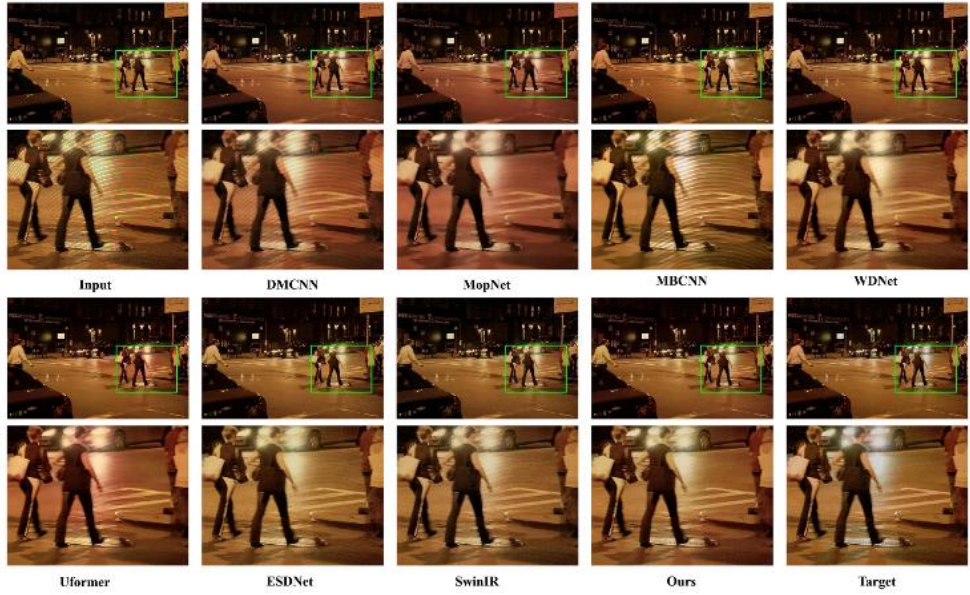

**Fig. 6** Subjective comparison results on SynMoiré.

The visualization results further substantiate that our SIDME model is more effective in preserving the structural and textural details of the image while concurrently reducing noise and artifacts, thereby exhibiting a clear superiority in subjective visual quality over other algorithms. These results underscore the effectiveness of our model's design, which incorporates a mask encoder-decoder reconstructor and a multidimensional loss function that includes self-supervised learning. By addressing the challenges of demoiréing from both synthetic and real-world datasets, our method not only delivers enhanced subjective visual quality but also achieves state-of-the-art performance in objective metrics. This comprehensive evaluation highlights the superiority of our approach over existing algorithms, establishing its robustness and applicability in practical scenarios.

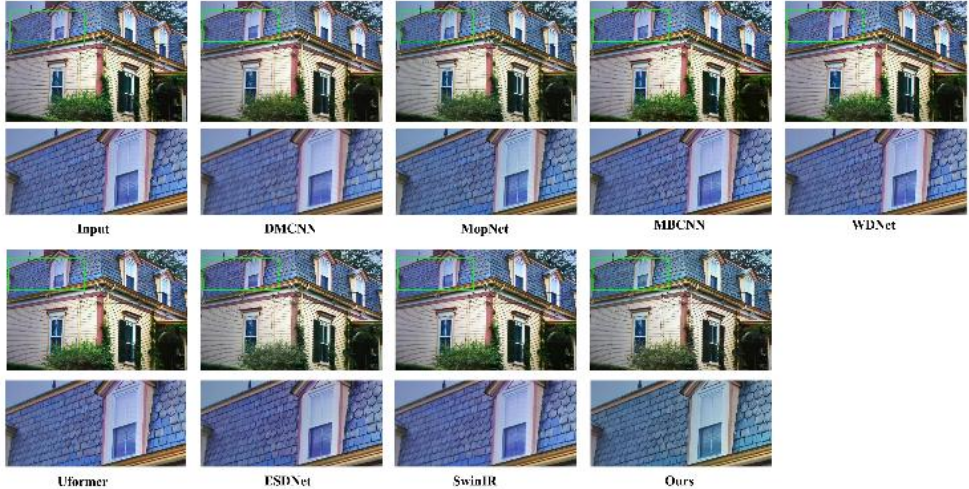

**Fig. 7** Subjective comparison results on real world data.

**Table 1** Evaluation Metrics (PSNR, SSIM, LPIPS) on SynMoiré.

|  | DMCNN | MopNet | MBCNN | WDNet | Uformer | ESDNet | SwinIR | Wang | P-BIC | Ours |
|---|---|---|---|---|---|---|---|---|---|---|
| PSNR | 28.89 | 30.93 | 28.81 | 28.83 | 29.67 | 32.65 | 32.16 | 31.15 | 32.98 | 34.94 |
| SSIM | 0.84 | 0.88 | 0.89 | 0.77 | 0.90 | 0.93 | 0.88 | 0.90 | 0.89 | 0.94 |
| LPIPS | 0.23 | 0.16 | 0.15 | 0.17 | 0.15 | 0.08 | 0.16 | 0.17 | 0.16 | 0.08 |

**Table 2** Evaluation Metrics (PSNR, SSIM, LPIPS) on TIP2018.

|  | DMCNN | MopNet | MBCNN | WDNet | Uformer | ESDNet | SwinIR | Wang | P-BIC | Ours |
|---|---|---|---|---|---|---|---|---|---|---|
| PSNR | 26.77 | 27.75 | 30.03 | 28.08 | 28.56 | 29.81 | 26.15 | 28.87 | 30.56 | 30.33 |
| SSIM | 0.87 | 0.90 | 0.89 | 0.90 | 0.91 | 0.92 | 0.87 | 0.98 | 0.93 | 0.92 |
| LPIPS | 0.20 | 0.177 | 0.177 | 0.17 | 0.18 | 0.16 | 0.20 | 0.18 | 0.16 | 0.14 |

## 5.3 Ablation Study

In this section, we experiment with different variants of the proposed method. Firstly, to identify the optimal mask ratio, we conduct model training and testing with various values, ranging from 0.15 to 0.85. As shown in Fig. 8, the model achieves the best performance when the mask ratio is set to 0.6, striking a balance between effective mask cover-age and computational efficiency. This can be attributed to the mask ratio of 0.6, which ensures that the model acquires sufficient contextual information (constituting 40% of the input) to guide the reconstruction process. Such contextual information is crucial for the model to accurately infer the missing portions. During training, the mask also serves as a regularizing agent, encouraging the model to learn generalized representations rather than overfitting to the visible data. Consequently, a mask ratio of 0.6 provides an appropriate level of regularization, ensuring that the model does not overfit while maintaining performance.

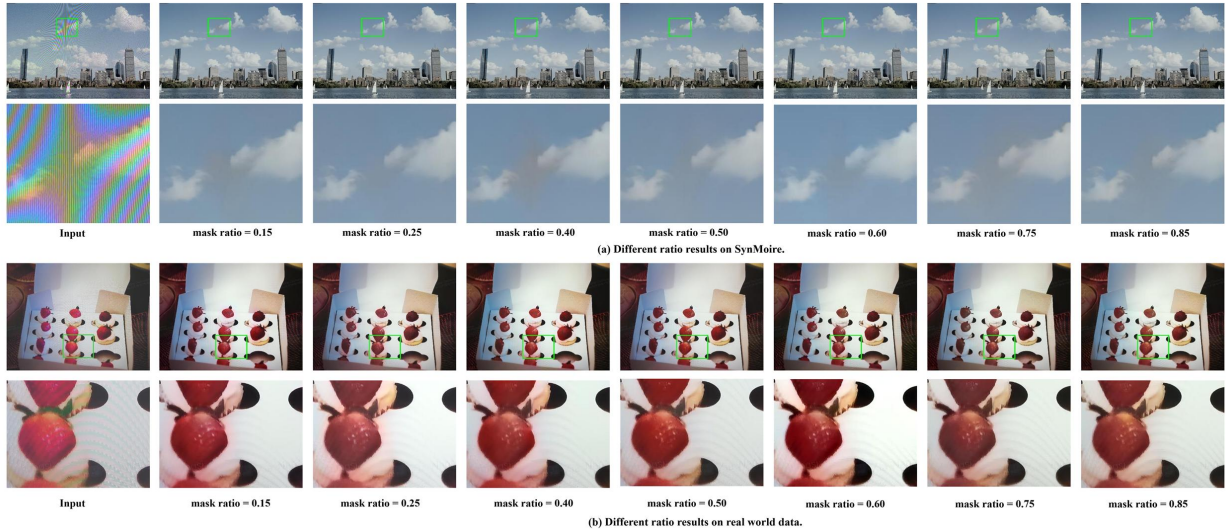

**Fig. 8** Results of different mask ratio.

Additionally, as depicted in Tab.3 and Fig. 9, we design targeted experiments to validate the effectiveness of the different components proposed. FA and ME denote feature aggregator and mask encoder-decoder, respectively. The results confirm that the synergy among these components leads to the overall best performance, outperforming each individual component and providing compelling evidence for the effectiveness of the pro-posed method.

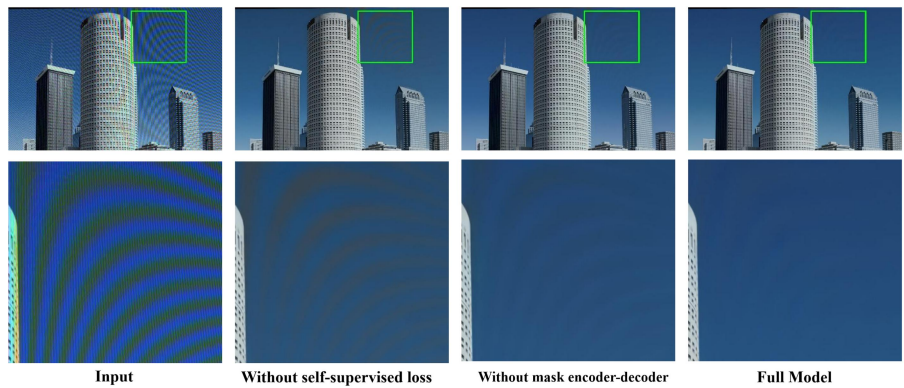

**Fig. 9** Experimental results of different variants of the Model on SynMoire.

Subsequently, we perform an ablation analysis for the proposed combination of loss functions, where B, S, P, E denotes basic loss, self-supervised loss, perceptual loss, and edge loss, respectively. The results in Tab.4 demonstrate the effectiveness of the multidimensional loss function in terms of performance improvement. The above ablation experiments not only demonstrate the critical role of each individual component but also highlight the importance of their interactions in achieving state-of-the-art results.

**Table 3** Evaluation Metrics (PSNR, SSIM, LPIPS) on SynMoiré.

|       | Basic (without-FA&ME) | Without_FA | Without_ME | All   |
|-------|----------------------|------------|------------|-------|
| PSNR  | 26.54                | 29.45      | 29.08      | 34.94 |
| SSIM  | 0.87                 | 0.90       | 0.89       | 0.94  |
| LPIPS | 0.18                 | 0.14       | 0.15       | 0.08  |

**Table 4** Ablation analysis of loss combinations on SynMoiré.

|       | B     | B+S   | B+P   | B+E   | B+S+P | B+S+E | B+E+P | Ours  |
|-------|-------|-------|-------|-------|-------|-------|-------|-------|
| PSNR  | 26.54 | 29.45 | 29.08 | 29.36 | 30.30 | 30.83 | 29.98 | 34.94 |
| SSIM  | 0.87  | 0.90  | 0.89  | 0.90  | 0.92  | 0.92  | 0.91  | 0.94  |
| LPIPS | 0.18  | 0.14  | 0.155 | 0.144 | 0.12  | 0.12  | 0.13  | 0.08  |

Subsequently, we perform an ablation analysis for the proposed combination of loss functions, where B, S, P, E denotes basic loss, self-supervised loss, perceptual loss, and edge loss, respectively. The results in Tab.4 demonstrate the effectiveness of the multidimensional loss function in terms of performance improvement.

The above ablation experiments not only demonstrate the critical role of each individual component but also highlight the importance of their interactions in achieving state-of-the-art results.

## 6  Conclusion

This paper is dedicated to addressing the problem of moiré pattern removal in image processing. Through research on the generation mechanism of moiré patterns and the characteristics of image sampling, we propose an image demoiréing method based on mask encoder-decoder reconstruction and self-supervised loss. This method fully exploits the intrinsic relationship between moiré patterns and image sampling and effectively enhances the generalization ability of the model through self-supervised learning. To vali-date the effectiveness of our method, we have constructed the SynMoiré dataset, which contains a wealth of moiré pattern image samples, providing robust data support for model training and testing. Experimental results demonstrate that our method not only removes moiré patterns but also preserves image details well, achieving superior performance compared to existing methods. Furthermore, through comparative experiments and ablation studies, we further confirm the pivotal roles of the mask encoder-decoder re-construction module and the self-supervised loss function in improving demoiréing performance.

*Code, Data, and Materials Availability*

The synthetic moiré dataset SynMoiré designed and implemented in this paper is publicly available at "https://ieee-dataport.org/documents/synmoire". The code for simulating and synthesizing moiré patterns and the model code designed and implemented in this paper can be obtained at "https://github.com/WangxiaBupt/SIDME".

**Xia Wang** received the B.S. degree in software engineering from the Shandong University of Science and Technology in 2021. She is currently pursuing the M.S. degree in computer technology at the Beijing University of Posts and Telecommunications. Her primary research focus is on image processing.

**Haiyang Sun** received the B.S. degree in software engineering from Shandong University of Science and Technology, in 2021. Received M. S. degree from the School of Artificial Intelligence, University of Chinese Academy of Sciences in 2024. He is currently pursuing the Ph. D degree at the School of Electronic Information and Electrical Engineering of Shanghai Jiao Tong University. His current research interests include deep learning and multimodal analysis.

**Tiantian Cao** received the B.S. degree in software engineering from the Beijing University of Posts and Telecommunications in 2022. She is currently pursuing the M.S. degree in artificial intelligence at the Beijing University of Posts and Telecommunications. Her primary research focus is on image processing.

**Yueying Sun** was born in China. She received the B.S. degree in software engineering from the Shandong University of Science and Technology in 2021. Received M.S. degree in computer science and technology from Shanghai Ocean University in 2024. Her primary research focus is on image and intelligent systems, with a concentration on Computer Vision and the application of Deep Learning technologies. She has authored four academic papers as the lead author and holds one invention patent. Her relevant achievements have been published in journals such as Fisheries Research and Transactions of the Chinese Society of Agricultural Engineering. Biographies and photographs for the other authors are not available.

**Min Feng** was born in China. She received the B.S. degree in information and computing science from the Shenyang University of Technology in 2021. Received M.S. degree in computer science and technology from Beijing University of Posts and Telecommunications in 2024. Her primary research focus is on Multimodal.

**Caption List**

**Fig. 1** Exemplary image of authentic moiré patterns.

**Fig. 2** Example of Bayer color filter array.

**Fig. 3** RGB channels of authentic moiré images.

**Fig. 4** Example of the Dataset.

**Fig. 5** he structure of the SIDME and the proposed feature aggregator (FA).

**Fig. 6** Subjective comparison results on SynMoiré.

**Fig. 7** Subjective comparison results on real world data.

**Fig. 8** Results of different mask ratio.

**Fig. 9** Experimental results of different variants of the Model on SynMoire.

**Table 1** Evaluation Metrics (PSNR, SSIM, LPIPS) on SynMoiré.

**Table 2** Evaluation Metrics (PSNR, SSIM, LPIPS) on TIP2018.

**Table 3** Ablation analysis of different components on SynMoiré.

**Table 4** Ablation analysis of loss combinations on SynMoiré.